\documentclass{sig-alternate-05-2015}
\usepackage{graphicx}
\usepackage{amsmath,amssymb} 
\usepackage{color}
\usepackage[table]{xcolor}
\usepackage[marginal]{footmisc}

\usepackage{url}
\usepackage{multirow}
\usepackage{subfigure}
\usepackage{algorithm}
\usepackage{algorithmic}

\begin{document}

\setcopyright{acmcopyright}

\doi{10.475/123_4}

\isbn{123-4567-24-567/08/06}

\conferenceinfo{PLDI '13}{June 16--19, 2013, Seattle, WA, USA}

\acmPrice{\$15.00}

%
\conferenceinfo{WOODSTOCK}{'97 El Paso, Texas USA}

\title{Content-based Video Relevance Prediction Challenge: Data, Protocol, and Baseline}
%
%
%
%
%

%
\author{
%
%
Mengyi Liu, Xiaohui Xie, Hanning Zhou\\
\\
\email{Hulu LLC., Beijing, 100120, China}\\
\email{mengyi.liu2012@gmail.com, \{mengyi.liu, xiaohui.xie, eric.zhou\}@hulu.com}
}

\maketitle
\begin{abstract}
Video relevance prediction is one of the most important tasks for online streaming service. Given the relevance of videos and viewer feedbacks, the system can provide personalized recommendations, which will help the user discover more content of interest. In most online service, the computation of video relevance table is based on users' implicit feedback, e.g. watch and search history. However, this kind of method performs poorly for ``cold-start'' problems - when a new video is added to the library, the recommendation system needs to bootstrap the video relevance score with very little user behavior known. One promising approach to solve it is analyzing video content itself, i.e. predicting video relevance by video frame, audio, subtitle and metadata. In this paper, we describe a challenge on Content-based Video Relevance Prediction (CBVRP) that is hosted by Hulu in the ACM Multimedia Conference 2018. In this challenge, Hulu drives the study on an open problem of exploiting content characteristics directly from original video for video relevance prediction. We provide massive video assets and ground truth relevance derived from our really system, to build up a common platform for algorithm development and performance evaluation.
\end{abstract}

%
%

%
%

%
%



\section{Introduction}

\label{section:secIntro}
\begin{figure}[t]
\centering
\includegraphics[height=4.6cm]{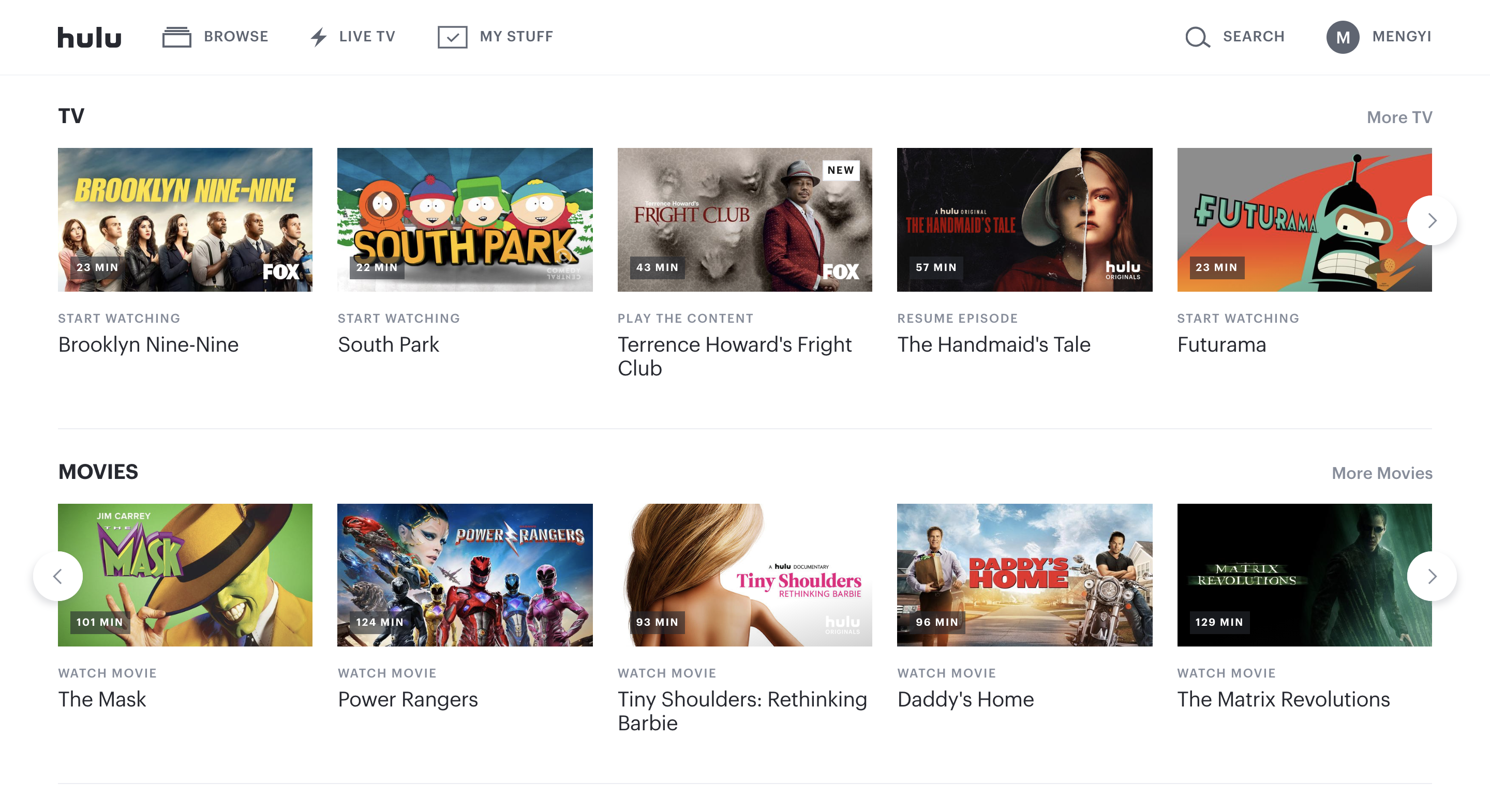}
\caption{Personalized recommendation cases in Hulu (for TV-shows and movies respectively).}
\label{figure:hulu}
\end{figure}

\vspace{5pt}
Video streaming service, such as YouTube, Netflix, and Hulu, depend heavily on the video recommendation system to help its user discover videos they would enjoy (see Figure~\ref{figure:hulu} for personalized recommendation cases in Hulu). Most existing recommendation systems compute the video relevance based on users' implicit feedback, e.g. watch and search behaviors. The system analyze the user-to-video preference and compute the video-to-video relevance scores using collaborative filtering based methods \cite{sarwar2001item,hu2008collaborative,wang2011collaborative,wang2015collaborative,zheng2016neural}. However, this kind of method performs poorly on ``cold-start'' problems - when a new video is added to the library, the recommendation system needs to bootstrap the video relevance score with very little user behavior known. One promising approach to solve ``cold-start'' is exploiting video content for relevance prediction, i.e. we can predict the video relevance by analyzing the content of videos including image pixels, audios, subtitles and metadata. Since the content contains almost all the information about a video, ideally, we can have enough details to build the video relevance table only from video content.

Generally, content-based methods focus on recommending items which have similar content characteristics to the items the user liked in the past. One of the key issues is how to extract the most relevant content features of each item. For most existing systems, the content features are associated with the items as structured metadata, e.g. movie/show genre, director/actors, description; Or other unstructured information from external sources, such as tags, and textual reviews. In contrast to these kinds of ``explicit'' features, there are also ``implicit'' content characteristics which can be exploited from the original movie/show video. Such characteristics could be visual features encoding low-level information like lighting, color, shape, motion, or high-level semantics like plot, mood, and artistic style.

To drive the study on this open problem, Hulu organized a challenge on Content-Based Video Relevance Prediction (CBVRP) to provide a common platform for algorithm development and evaluation. In this paper, we will introduce the detailed information of the challenge task, including problem formulation and the dataset with evaluation protocol. We also provide baseline results on validation and test set respectively, which can be referred by participants and other researchers.

\section{Data and Protocol}

\vspace{5pt}
The main task of this challenge is to predict the relevance between TV-shows or movies from video content and its features. Due to legal and copyright issue, we utilize TV-show or movie trailers instead of full length videos as content data. Specifically, there are two separated tracks for TV-shows and movies respectively \footnote{The dataset is available at: https://github.com/mengyi-liu/cbvrp-acmmm-2018}.

\textbf{Track 1 - TV shows}: For TV-shows, we provide nearly 7,000 TV-show video trailers with pre-extracted features. The whole set is divided into 3 subsets: training set (3,000 shows), validation set (864 shows), and testing set (3,000 shows). All of the TV-shows are categorized into 35 distinct genres. Table~\ref{tab:showGenre} illustrates the top-10 high frequency ones with their name and occurrence.

\begin{table}[tbh]
\linespread{1.2}
\small
\caption{Top-10 high frequency genres for shows.}
\centering
\vspace{5pt}
\begin{tabular}{l|c}
\hline
Genre & Occurence \\
\hline\hline
Drama & 1216 \\
Animation & 1135 \\
Reality and Game Show & 1003 \\
Comedy & 867 \\
Documentary and Biography & 707 \\
Family and Kids & 375 \\
Lifestyle and Fashion & 349 \\
News and Information & 233 \\
Mixed Martial Arts/Fighting & 216 \\
Action and Adventure & 204 \\
\hline\hline
\end{tabular}
\label{tab:showGenre}
\end{table}

\textbf{Track 2 - Movies}: For movies, we provide over 10,000 movie video trailers. The whole set is divided into 3 subsets: training set (4,500 movies), validation set (1188 movies), and testing set (4,500 movies). For movies, there are 27 distinct genres. Table~\ref{tab:movieGenre} illustrates the top-10 high frequency ones for movies with their name and occurrence.

\begin{table}[tbh]
\linespread{1.2}
\small
\caption{Top-10 high frequency genres for movies.}
\centering
\vspace{5pt}
\begin{tabular}{l|c}
\hline
Genre & Occurence \\
\hline\hline
Drama & 2730 \\
Documentary and Biography & 2286 \\
Comedy & 1798 \\
Action and Adventure & 1174 \\
Horror and Thriller & 1160 \\
Family and Kids & 571 \\
Animation & 317 \\
Crime and Mystery & 186 \\
Sci Fi and Fantasy & 180 \\
Music & 115 \\
\hline\hline
\end{tabular}
\label{tab:movieGenre}
\end{table}

For training set and validation set in both tracks, we also provide the ground truth (relevance lists) derived from implicit viewer feedbacks. The viewer feedbacks have been cleaned to avoid any privacy issues. Specifically, for each item $r$, we provide the ground truth top $M$ most relevant items retrieved from candidate set $C$. The relevance list is defined as $o^r = [o^r_1, o^r_2, ..., o^r_M]$, where $o^r_i \in C$ is the item ranked $i$-th position in $o^r$. The participants need to predict top $K$ relevant shows/movies for each item, which can be represented as $\widetilde{o^r} = [\widetilde{o^r_1}, \widetilde{o^r_2}, ..., \widetilde{o^r_K}]$. The submission results will be evaluated based on $recall$ and $hit$ rate regarding to top K prediction. Based on above formulation, the metric $recall@K$ can be calculated as following:

\begin{equation}
recall@K = \frac{|o^r \bigcap \widetilde{o^r}|}{|o^r|}.
\end{equation}

While $hit@K$ is defined for as single test case as either value 1 if $recall@K > 0$, or the value 0 if otherwise. In the end, we report the average of $recall$ and $hit$ on all test cases respectively as the final performance. For fair comparison, we also provide the python script to compute the metrics for evaluation.

\section{Baseline Method}

\vspace{5pt}
In this section, we introduce our baseline method for content based video relevance prediction, which can be referred by participants and other researchers.

\subsection{Feature Representation}

\vspace{5pt}
Based on recent advances in computer vision, more and more deep convolutional neural networks are developed that become increasingly proficient in mimicking perceptual inference abilities of humans. As a side effect of their popularity in technology, the increasing availability and diversity of high-performing neural network models opens a new door for studying the neural mechanisms of perceptual skills such as object recognition and scene understanding. Therefore, in this challenge, we utilize deep neural network features for the video representation. Specifically, for \textbf{frame-level} features, we decode each video at 1 fps and then feed the decoded frames into the InceptionV3 networks \cite{abu2016youtube} trained on ImageNet dataset, and fetch the ReLU activation of the last hidden layer, named $pool\_3/\_reshape$ with 2048 dimensions. After obtaining frame features, we further compute their mean representation as the final signature. For \textbf{video-level} features, we also employ the state-of-the-art architectures - C3D models, resorting to its most popular implementation from Facebook \cite{tran2015learning} respectively. Here we decode each video at 8 fps and feed the frame stream into the model trained on Sports1M dataset \cite{karpathy2014large}, and fetch the activations of pool5 layer with 512 dimensions as the final video clip feature.

\begin{table*}[t]
\linespread{1.5}
\small
\caption{Track 1 - TV-shows: performance comparison of different parameters for triplet network training (Based on C3D layer-pool5 feature): (a) Validation set. (b) Testing set.}
\centering
\vspace{5pt}
\subtable[Validation set]{
\begin{tabular}{c|c|cccc|cccc}
\hline
\multirow{2}{*}{\#dim} & \multirow{2}{*}{\#epoch} & \multicolumn{4}{c|}{hit@k} & \multicolumn{4}{c}{recall@k} \\
\cline{3-10}
 & & ~~k=5~~ & ~k=10~ & ~k=20~ & ~k=30~ & ~~k=50~ & ~k=100~ & ~k=200~ & ~k=300~ \\
\hline\hline
\multirow{3}{*}{64} & 4 & 0.244 & 0.326 & 0.431 & \textbf{0.510} & 0.109 & 0.172 & \textbf{0.264} & \textbf{0.329} \\
                    & 8 & 0.247 & 0.329 & 0.439 & 0.497 & \textbf{0.111} & 0.173 & 0.263 & \textbf{0.329} \\
                    & 16 & 0.241 & 0.325 & 0.425 & 0.481 & 0.108 & \textbf{0.175} & 0.261 & 0.326 \\
\hline
\multirow{3}{*}{128} & 4 & 0.237 & \textbf{0.347} & \textbf{0.442} & 0.505 & 0.109 & 0.171 & 0.261 & 0.325 \\
                     & 8 & \textbf{0.253} & 0.338 & 0.427 & 0.501 & 0.110 & 0.172 & 0.260 & 0.327 \\
                     & 16 & 0.245 & 0.336 & 0.426 & 0.493 & 0.108 & 0.172 & 0.260 & 0.326 \\
\hline
\multirow{3}{*}{256} & 4 & 0.237 & 0.322 & 0.436 & 0.499 & 0.108 & 0.169 & 0.255 & 0.323 \\
                     & 8 & 0.243 & 0.325 & 0.434 & 0.499 & 0.110 & 0.170 & 0.256 & 0.323 \\
                     & 16 & 0.235 & 0.317 & 0.432 & 0.491 & 0.108 & 0.166 & 0.251 & 0.318 \\
\hline
\hline
\end{tabular}
}
\subtable[Testing set]{
\begin{tabular}{c|c|cccc|cccc}
\hline
\multirow{2}{*}{\#dim} & \multirow{2}{*}{\#epoch} & \multicolumn{4}{c|}{hit@k} & \multicolumn{4}{c}{recall@k} \\
\cline{3-10}
 & & ~~k=5~~ & ~k=10~ & ~k=20~ & ~k=30~ & ~~k=50~ & ~k=100~ & ~k=200~ & ~k=300~ \\
\hline\hline
\multirow{3}{*}{64} & 4 & 0.215 & 0.314 & 0.427 & 0.492 & 0.076 & 0.130 & 0.205 & \textbf{0.257} \\
                    & 8 & 0.212 & 0.306 & 0.423 & 0.490 & 0.075 & 0.129 & 0.202 & 0.252 \\
                    & 16 & 0.212 & 0.307 & 0.416 & 0.485 & 0.074 & 0.127 & 0.198 & 0.248 \\
\hline
\multirow{3}{*}{128} & 4 & 0.225 & 0.322 & 0.430 & 0.498 & 0.077 & 0.130 & \textbf{0.206} & 0.256 \\
                     & 8 & 0.225 & 0.309 & 0.422 & 0.493 & 0.076 & 0.129 & 0.202 & 0.251 \\
                     & 16 & 0.219 & 0.313 & 0.415 & 0.487 & 0.075 & 0.126 & 0.197 & 0.247 \\
\hline
\multirow{3}{*}{256} & 4 & 0.231 & \textbf{0.328} & 0.442 & \textbf{0.510} & \textbf{0.079} & \textbf{0.132} & \textbf{0.206} & 0.255 \\
                     & 8 & \textbf{0.234} & 0.327 & \textbf{0.444} & 0.509 & 0.077 & 0.131 & 0.205 & 0.254 \\
                     & 16 & 0.232 & 0.321 & 0.426 & 0.497 & 0.076 & 0.128 & 0.199 & 0.247 \\
\hline
\hline
\end{tabular}
}
\label{table:shows}
\end{table*}

\subsection{Relevance Learning}

\vspace{5pt}
According to the ground truth given, we cannot obtain the ``categories'' of videos, the only supervision is the relationship between two videos, that is, relevant or not relevant. To deal with such kind of pair-wise constraints, many recent approaches have employed ``triplet loss'' or its variants for the model training \cite{schroff2015facenet,shi2016embedding,liu2016multi,cheng2016person,liu2017end,hermans2017defense}, and also achieved significant success. In our baseline method, we utilize the basic triplet loss for relevance learning. For triplet construction, we define each query as the ``anchor'', the items appeared in its relevance list are considered as ``positive'' samples, while the others are considered as ``negative'' samples. Specifically, we denote the original feature for each show/movie trailer as $x$. The embedding function learning from our neural networks is represented by $f(x)\in R^d$, which embeds $x$ into a $d$-dimensional space. Given a triplet $(x^a, x^p, x^n)$, where $x^a$, $x^p$, $x^n$ are anchor, positive, and negative respectively. Since we aim to minimize the distance between the relevant (positive) pair while maximize the distance between irrelevant (negative) pair, the ``triplet loss'' can be represented as:

\begin{equation}
L_{tri} = \Sigma_{a, p, n}[m + D(f(x^a), f(x^p)) - D(f(x^a), f(x^n))]_+,
\end{equation}

where $m$ is a margin that is enforced between positive and negative pairs. The loss is optimized over the set of possible triplets constructed in the training set. In our experiments, we randomly select equal number of positive and negative samples for each anchor. The value of margin is fixed as $m = 1.0$ for all settings.

\section{Experiments}

\vspace{5pt}
In our experiments, we construct a triplet network with a single full-connection layer for relevance learning. The embedding dimension $d$ is tunable parameter varying in $64$, $128$, and $256$. We also tune the number of training epoch $\#epoch$ to evaluate the fitting process. As for validation purpose only, we only conduct such experiments based on \textbf{C3D-pool5} feature due to its lower dimension of 512. The detailed results based on different parameters are listed in Table~\ref{table:shows} and Table~\ref{table:movies}.

\begin{table*}[t]
\linespread{1.5}
\small
\caption{Track 2 - Movies: performance comparison of different parameters for triplet network training (Based on C3D layer-pool5 feature): (a) Validation set. (b) Testing set.}
\centering
\vspace{5pt}
\subtable[Validation set]{
\begin{tabular}{c|c|cccc|cccc}
\hline
\multirow{2}{*}{\#dim} & \multirow{2}{*}{\#epoch} & \multicolumn{4}{c|}{hit@k} & \multicolumn{4}{c}{recall@k} \\
\cline{3-10}
 & & ~~k=5~~ & ~k=10~ & ~k=20~ & ~k=30~ & ~~k=50~ & ~k=100~ & ~k=200~ & ~k=300~ \\
\hline\hline
\multirow{3}{*}{64} & 4 & 0.137 & 0.194 & 0.273 & 0.350 & 0.095 & 0.135 & 0.197 & 0.245 \\
                    & 8 & 0.140 & 0.197 & 0.281 & 0.337 & 0.094 & 0.135 & 0.198 & 0.245 \\
                    & 16 & 0.138 & 0.192 & 0.279 & 0.335 & 0.092 & 0.135 & 0.193 & 0.240 \\
\hline
\multirow{3}{*}{128} & 4 & 0.146 & 0.210 & 0.287 & 0.349 & 0.098 & 0.140 & 0.195 & 0.244 \\
                     & 8 & 0.154 & 0.209 & 0.298 & 0.354 & 0.098 & 0.138 & 0.200 & 0.244 \\
                     & 16 & 0.138 & 0.204 & \textbf{0.300} & 0.364 & 0.096 & 0.137 & 0.197 & 0.247 \\
\hline
\multirow{3}{*}{256} & 4 & \textbf{0.167} & 0.209 & 0.294 & 0.358 & \textbf{0.101} & 0.141 & 0.200 & 0.250 \\
                     & 8 & 0.158 & \textbf{0.213} & 0.294 & \textbf{0.366} & \textbf{0.101} & \textbf{0.143} & 0.205 & 0.255 \\
                     & 16 & 0.150 & 0.210 & 0.297 & 0.354 & 0.099 & 0.141 & \textbf{0.206} & \textbf{0.257} \\
\hline
\hline
\end{tabular}
}
\subtable[Testing set]{
\begin{tabular}{c|c|cccc|cccc}
\hline
\multirow{2}{*}{\#dim} & \multirow{2}{*}{\#epoch} & \multicolumn{4}{c|}{hit@k} & \multicolumn{4}{c}{recall@k} \\
\cline{3-10}
 & & ~~k=5~~ & ~k=10~ & ~k=20~ & ~k=30~ & ~~k=50~ & ~k=100~ & ~k=200~ & ~k=300~ \\
\hline\hline
\multirow{3}{*}{64} & 4 & 0.155 & 0.214 & 0.292 & 0.349 & 0.069 & 0.100 & 0.148 & 0.185 \\
                    & 8 & 0.158 & 0.214 & 0.292 & 0.350 & 0.068 & 0.098 & 0.144 & 0.180 \\
                    & 16 & 0.158 & 0.213 & 0.285 & 0.339 & 0.066 & 0.096 & 0.141 & 0.176 \\
\hline
\multirow{3}{*}{128} & 4 & 0.160 & 0.221 & 0.293 & 0.343 & 0.070 & 0.101 & 0.147 & 0.182 \\
                     & 8 & 0.163 & 0.218 & 0.290 & 0.346 & 0.068 & 0.099 & 0.143 & 0.179 \\
                     & 16 & 0.164 & 0.215 & 0.285 & 0.339 & 0.068 & 0.098 & 0.142 & 0.177 \\
\hline
\multirow{3}{*}{256} & 4 & \textbf{0.167} & \textbf{0.227} & \textbf{0.303} & \textbf{0.356} & \textbf{0.073} & \textbf{0.106} & \textbf{0.152} & \textbf{0.189} \\
                     & 8 & \textbf{0.167} & 0.218 & 0.300 & 0.355 & 0.072 & 0.104 & 0.151 & \textbf{0.189} \\
                     & 16 & 0.162 & 0.217 & 0.292 & 0.351 & 0.069 & 0.101 & 0.147 & 0.184 \\
\hline
\hline
\end{tabular}
}
\label{table:movies}
\end{table*}

\begin{table*}[t]
\linespread{1.5}
\small
\caption{Performance comparison between unsupervised cosine similarity and relevance learning via triplet network. (a) Track 1 - TV-shows. (b) Track 2 - Movies.}
\centering
\vspace{5pt}
\subtable[Track 1 - TV-shows]{
\begin{tabular}{c|l|cccc|cccc}
\hline
\multirow{2}{*}{} & \multirow{2}{*}{Feature} & \multicolumn{4}{c|}{hit@k} & \multicolumn{4}{c}{recall@k} \\
\cline{3-10}
 & & ~~k=5~~ & ~k=10~ & ~k=20~ & ~k=30~ & ~~k=50~ & ~k=100~ & ~k=200~ & ~k=300~ \\
\hline\hline
\multirow{4}{*}{Val} & Inception-pool3 & 0.234 & 0.316 & 0.397 & 0.448 & 0.083 & 0.124 & 0.192 & 0.244 \\
                     & Inception-pool3 (Tri) & 0.247 & 0.338 & 0.450 & 0.530 & 0.111 & 0.172 & 0.262 & 0.331 \\
                     \cline{2-10}
                     & C3D-pool5 & 0.234 & 0.313 & 0.409 & 0.488 & 0.092 & 0.145 & 0.216 & 0.267 \\
                     & C3D-pool5 (Tri) & 0.253 & 0.347 & 0.442 & 0.510 & 0.111 & 0.175 & 0.264 & 0.329 \\
                     \cline{2-10}
                     & Late fusion & 0.272 & 0.376 & 0.476 & \textbf{0.541} & 0.123 & \textbf{0.189} & 0.281 & 0.352 \\
\hline
\multirow{4}{*}{Test} & Inception-pool3 & 0.246 & 0.332 & 0.432 & 0.497 & 0.073 & 0.114 & 0.170 & 0.209 \\
                      & Inception-pool3 (Tri) & 0.209 & 0.301 & 0.412 & 0.485 & 0.075 & 0.123 & 0.196 & 0.244 \\
                      \cline{2-10}
                      & C3D-pool5 & 0.247 & 0.337 & 0.444 & 0.505 & 0.076 & 0.123 & 0.186 & 0.229 \\
                      & C3D-pool5 (Tri) & 0.234 & 0.328 & 0.444 & 0.510 & 0.079 & 0.132 & 0.206 & 0.257 \\
                      \cline{2-10}
                      & Late fusion & 0.249 & 0.356 & 0.461 & \textbf{0.525} & 0.085 & \textbf{0.141} & 0.219 & 0.269 \\
\hline\hline
\end{tabular}
}
\subtable[Track 2 - Movies]{
\begin{tabular}{c|l|cccc|cccc}
\hline
\multirow{2}{*}{} & \multirow{2}{*}{Feature} & \multicolumn{4}{c|}{hit@k} & \multicolumn{4}{c}{recall@k} \\
\cline{3-10}
 & & ~~k=5~~ & ~k=10~ & ~k=20~ & ~k=30~ & ~~k=50~ & ~k=100~ & ~k=200~ & ~k=300~ \\
\hline\hline
\multirow{4}{*}{Val} & Inception-pool3 & 0.141 & 0.185 & 0.248 & 0.291 & 0.072 & 0.099 & 0.137 & 0.167 \\
                     & Inception-pool3 (Tri) & 0.133 & 0.187 & 0.269 & 0.312 & 0.086 & 0.125 & 0.185 & 0.229 \\
                     \cline{2-10}
                     & C3D-pool5 & 0.140 & 0.193 & 0.271 & 0.316 & 0.084 & 0.112 & 0.160 & 0.196 \\
                     & C3D-pool5 (Tri) & 0.167 & 0.213 & 0.300 & 0.366 & 0.101 & 0.143 & 0.206 & 0.257 \\
                     \cline{2-10}
                     & Late fusion & 0.168 & 0.234 & 0.320 & \textbf{0.371} & 0.107 & \textbf{0.154} & 0.219 & 0.269 \\
\hline
\multirow{4}{*}{Test} & Inception-pool3 & 0.167 & 0.212 & 0.273 & 0.320 & 0.064 & 0.087 & 0.120 & 0.144 \\
                      & Inception-pool3 (Tri) & 0.155 & 0.207 & 0.276 & 0.325 & 0.067 & 0.096 & 0.139 & 0.173 \\
                      \cline{2-10}
                      & C3D-pool5 & 0.168 & 0.219 & 0.289 & 0.336 & 0.069 & 0.093 & 0.129 & 0.158 \\
                      & C3D-pool5 (Tri) & 0.167 & 0.227 & 0.303 & 0.356 & 0.073 & 0.106 & 0.152 & 0.189 \\
                      \cline{2-10}
                      & Late fusion & 0.190 & 0.242 & 0.320 & \textbf{0.373} & 0.081 & \textbf{0.116} & 0.168 & 0.206 \\
\hline\hline
\end{tabular}
}
\label{table:Base}
\end{table*}

To verify the effectiveness of relevance learning via triplet network, we also generate the baseline results in fully unsupervised setting, that is, calculating cosine similarity between query and candidate items for relevance ranking. Two kinds of features, \textbf{Inception-pool3} and \textbf{C3D-pool5} are evaluated on validation and testing set respectively. For triplet network, we set the embedding dimension $d = 256$ for both features, which are denoted as \textbf{Inception-pool3 (Tri)} and \textbf{C3D-pool5 (Tri)}. To further improve the performance, we also perform late fusion on these two features (with $d = 256$ and $\#epoch = 4$) by averaging the two similarity matrices. All of the results for baseline and triplet network are illustrated in Table~\ref{table:Base}. We can observe the improvement brought by triplet network on both of the features (especially with larger value of $k$). And the late fusion among two features also boost the final performance.

\section{Conclusions}

\vspace{5pt}
In this paper, we introduce the detailed information of the Content based Video Relevance Prediction (CBVRP) challenge task, including problem formulation and the dataset with evaluation protocol. Baseline results are also provided on validation and test set respectively for reference. In the future work, we will attempt to involve multi-modal content features, including visual, audio, and textual channels, and further investigate more effective algorithms for relevance learning. This challenge provides good opportunities for both organizers and participants to better understand this problem with real-world data and scenario. We will continue to work with the community to explore and improve the understanding on video content analysis.

\bibliographystyle{splncs}
\bibliography{egbib}
\end{document}